\documentclass[a4paper]{spie}  
\usepackage[]{graphicx,epsfig,url}
\title{Automated eye disease classification method from anterior eye image using anatomical structure focused image classification technique}

\author{Masahiro ODA\supit{a}, Takefumi YAMAGUCHI\supit{b}, Hideki FUKUOKA\supit{c},\\ Yuta UENO\supit{d}, and Kensaku MORI\supit{a,e}
\skiplinehalf
\supit{a}Graduate School of Informatics, Nagoya University, \\
Furo-cho, Chikusa-ku, Nagoya, Aichi, 464-8603, Japan; \\
\supit{b}Department of Ophthalmology, Tokyo Dental College, Ichikawa General Hospital, \\
5-11-13, Sugano, Ichikawa-shi, Chiba, 272-8513, Japan; \\
\supit{c}Department of Ophthalmology, Kyoto Prefectural University of Medicine, \\
Kawaramachi-Hirokoji, Kajii-cho, Kamigyo-ku, Kyoto-shi, Kyoto, 602-8566, Japan; \\
\supit{d}Department of Ophthalmology, Faculty of Medicine, University of Tsukuba, \\
1-1-1, Tennodai, Tsukuba-shi, Ibaraki, 305-8575, Japan; \\
\supit{e}Research Center for Medical Bigdata, National Institute of Informatics, \\
2-1-2, Hitotsubashi, Chiyoda-ku, Tokyo, 101-8430, Japan;
}

\authorinfo{Further author information: 
(Send correspondence to M. Oda)
\\M. Oda: E-mail: moda@mori.m.is.nagoya-u.ac.jp, Telephone: +81 (0)52 789 5688
\\  K.Mori: E-mail: kensaku@is.nagoya-u.ac.jp, Telephone: +81 (0)52 789 5689
}

\begin{document} 
  \maketitle 

\begin{abstract}
This paper presents an automated classification method of infective and non-infective diseases from anterior eye images.
Treatments for cases of infective and non-infective diseases are different.
Distinguishing them from anterior eye images is important to decide a treatment plan.
Ophthalmologists distinguish them empirically.
Quantitative classification of them based on computer assistance is necessary.
We propose an automated classification method of anterior eye images into cases of infective or non-infective disease.
Anterior eye images have large variations of the eye position and brightness of illumination.
This makes the classification difficult.
If we focus on the cornea, positions of opacified areas in the corneas are different between cases of the infective and non-infective diseases.
Therefore, we solve the anterior eye image classification task by using an object detection approach targeting the cornea.
This approach can be said as ``anatomical structure focused image classification''.
We use the YOLOv3 object detection method to detect corneas of infective disease and corneas of non-infective disease.
The detection result is used to define a classification result of a image.
In our experiments using anterior eye images, 88.3\% of images were correctly classified by the proposed method.
\end{abstract}



\section{Introduction}

Diseases of the eye can be roughly classified into infective and non-infective diseases.
Eye infection occurs by bacteria, fungi, and viruses \cite{Watson18}.
This causes red eyes, pain, itching, and blurry vision.
Non-infective diseases include ulcers and trauma.
Treatments for infective and non-infective diseases cases are different.
Distinguish of infective disease or non-infective one is an important task in diagnosis.
Ophthalmologists distinguish them by observing the anterior eye.
However, the criteria for distinguish the two diseases are made based on experience of individual ophthalmologist.
Decisions made by a less experienced ophthalmologist have possibility to be wrong.
Decision assistance by a computer is necessary to provide diagnoses at a certain level.

\begin{figure}[tb]
\begin{center}
\begin{tabular}{cc}
\includegraphics[width=0.45\textwidth, clip, trim=0 340 620 0]{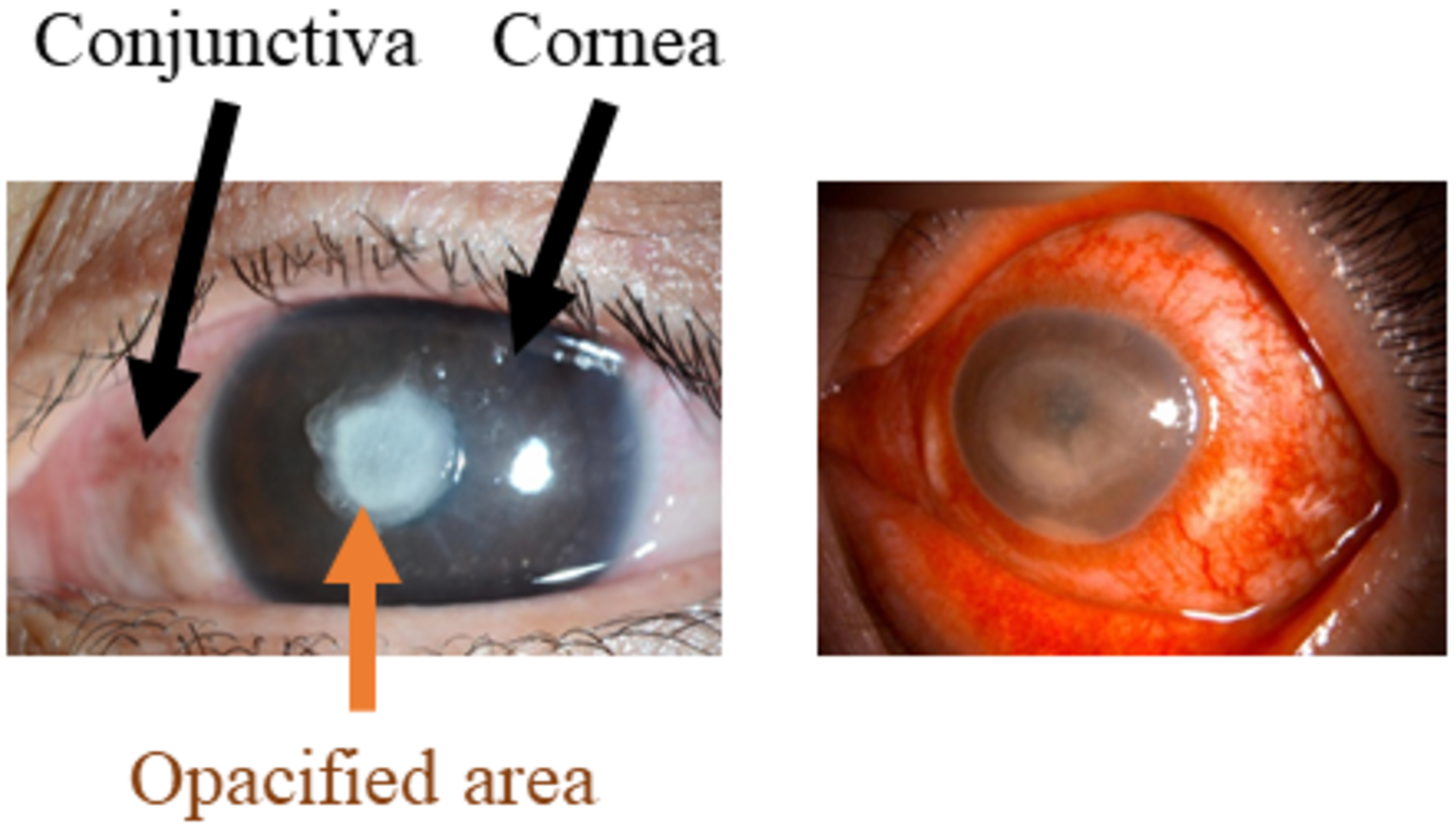} & 
\includegraphics[width=0.45\textwidth, clip, trim=0 340 620 0]{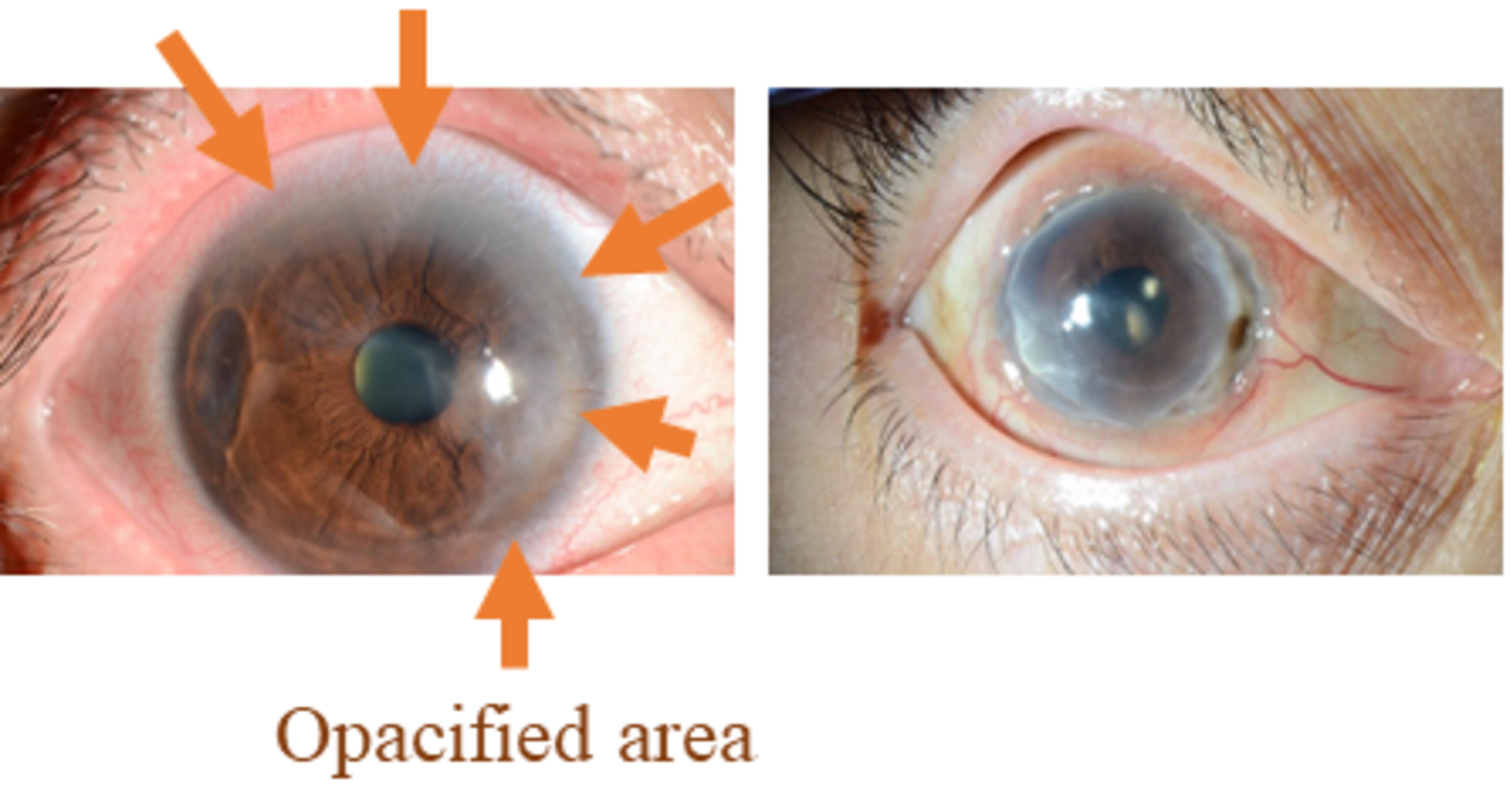} \\
(a) & (b)
\end{tabular}
\end{center}
\caption{Typical anterior eye images of cases of (a) infective and (b) non-infective diseases. Positions of opacified areas in the cornea are different among them.}
\label{fig:sample_opacified}
\end{figure}

\begin{figure}[tb]
\begin{center}
\begin{tabular}{c}
\includegraphics[width=0.95\textwidth, clip, trim=0 460 270 0]{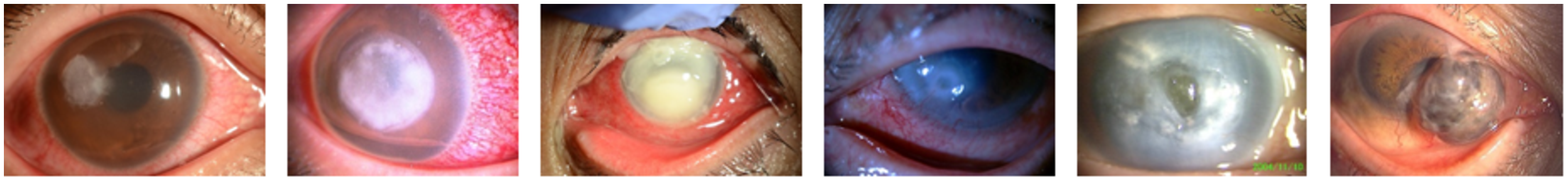} \\
(a) \\
\includegraphics[width=0.95\textwidth, clip, trim=0 460 270 0]{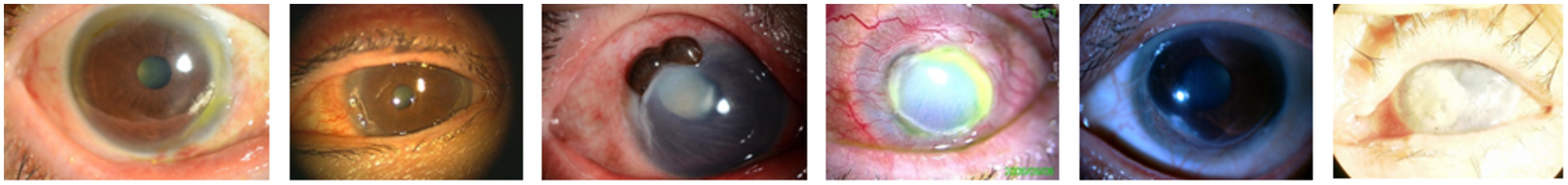} \\
(b)
\end{tabular}
\end{center}
\caption{Anterior eye image samples of cases of (a) infective and (b) non-infective diseases.}
\label{fig:sample}
\end{figure}

The cornea and conjunctiva can be observed in anterior eye images.
Typical anterior eye images of cases of infective and non-infective diseases are shown in Fig. \ref{fig:sample_opacified}.
Characteristics of the eye with infective disease are: center of the cornea is opacified and redness of the conjunctiva is observed.
Characteristics of the eye with non-infective disease are: marginal area of the cornea is opacified and redness of the conjunctiva varies among patients.
More samples of anterior eye images of infective and non-infective diseases are shown in Fig. \ref{fig:sample}.
As shown in this figure, not only the position of opacified area and redness of the conjunctiva, but also the eye position, degree of open of the eyelid, and brightness of illumination greatly varies among the images.
These variations make automated diagnosis difficult.
Automated diagnosis assistance methods for infective and non-infective disease from anterior eye images have not been developed.

Recently, deep learning techniques are used to develop automated diagnosis assistances of the eye.
However, most of such methods process fundus images to perform anatomical structure segmentations and lesion detections \cite{Fu16,Wu16,Liskowski16,Prentasic16,Dasgupta17,Zhang18,Wu18}.
Despite anterior eye images contain useful information for diagnosis, they are not commonly utilized in automated processes for diagnosis assistance.

In this paper, we propose an automated classification method of cases of infective and non-infective diseases from anterior eye images.
To the best of our knowledge, this is the first automated method for infective and non-infective diseases diagnosis assistance from anterior eye images.
As we explained above, there are many variations of appearances among anterior eye images that make classification difficult.
We need to focus on a specific anatomical structure that shows different appearances between cases of infective and non-infective diseases.
Appearances of the cornea and conjunctiva are useful to differentiate cases of infective and non-infective diseases.
Especially, the position of opacified area in the cornea is different among the two disease cases.
Therefore, we solve the image classification task by using an object detection approach targeting the cornea.
Our approach can be said as ``anatomical structure focused image classification''.
We use an object detection approach not for object detection but also classification of images.

We use the YOLOv3 \cite{yolo3} object detection method.
The YOLOv3 is trained to detect two targets including the cornea of infective disease and the cornea of non-infective disease from anterior eye images.
In the inference stage, we classify an anterior eye image into one of infective or non-infective disease class based on the detection result of the YOLOv3.
This process enables image classification based on difference of appearances of the cornea, which is important for diagnosing infective and non-infective diseases.

\section{Method} \label{sec:method}

\subsection{Training data}

\begin{figure}[tb]
\begin{center}
\begin{tabular}{c}
\includegraphics[width=0.75\textwidth, clip, trim=0 420 400 0]{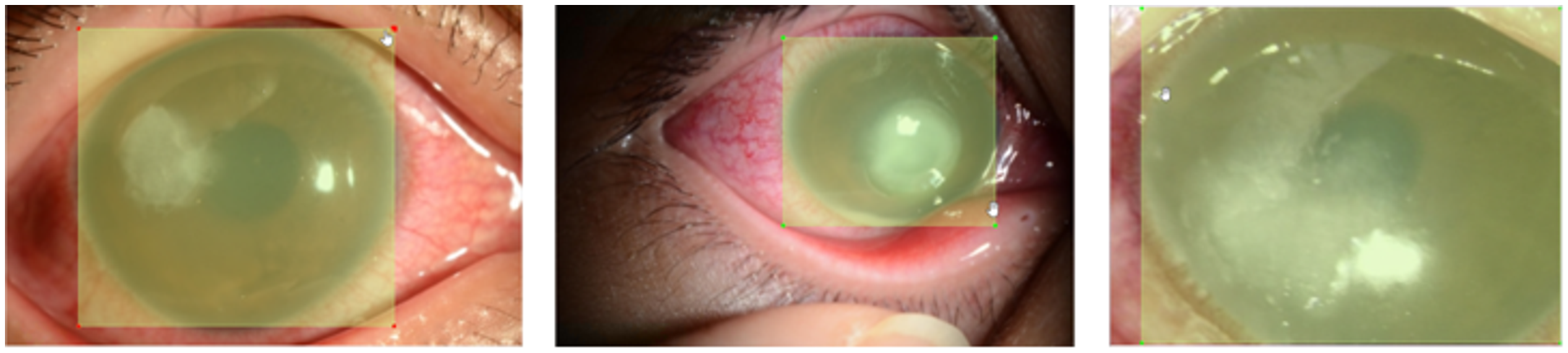} \\
(a) \\
\includegraphics[width=0.75\textwidth, clip, trim=0 420 400 0]{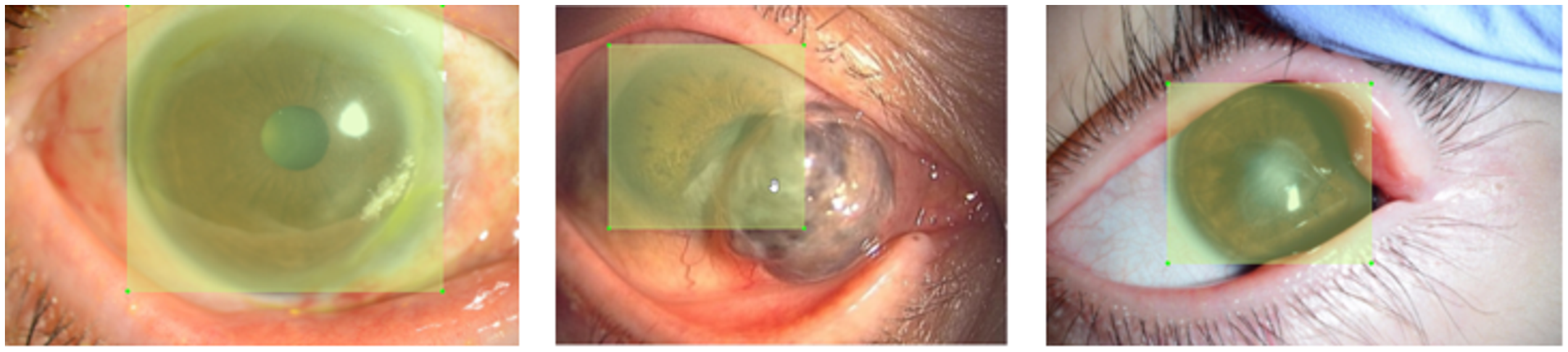} \\
(b)
\end{tabular}
\end{center}
\caption{Samples of annotations for (a) infective and (b) non-infective disease images. Yellow boxes are bounding boxes with ``infective cornea'' or ``non-infective cornea'' label.}
\label{fig:annotation}
\end{figure}

The training data contains anterior eye images taken from infective and non-infective disease patients.
We manually provide annotations to the images.
For the images of the infective disease patients, ``infective cornea'' label and bounding boxes covering the corneas are given as annotations.
For the images of the non-infective disease patients, ``non-infective cornea'' label and bounding boxes covering the corneas are given as annotations.
Samples of annotated images are shown in Fig. \ref{fig:annotation}.
The images and corresponding annotations are used to train the object detection method.

%
%

\subsection{Anatomical structure focused anterior eye image classification method}

In the YOLO framework \cite{yolo1}, a convolutional neural network (CNN) is used to perform estimation of bounding boxes, object class labels, and objectness scores from an input image.
Multiple bounding boxes can be detected from an input image.
We train the YOLOv3 using the images for training and corresponding annotations.
The object classes include ``infective cornea'' and ``non-infective cornea''.
After the training, we input images for testing to the trained YOLOv3.
We obtain estimated bounding boxes, object class labels, and objectness scores of the images.

\begin{figure}[tb]
\begin{center}
\begin{tabular}{c}
\includegraphics[width=0.5\textwidth, clip, trim=0 350 470 0]{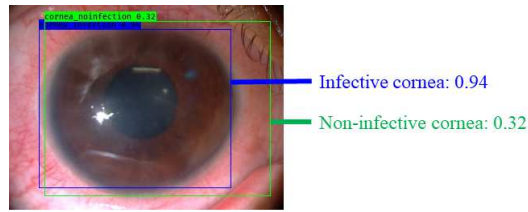} 
\end{tabular}
\end{center}
\caption{Example of non-maximal objectness score removal. In estimation results, multiple bounding boxes can be contained. Boxes in figure are samples of detected bounding boxes. Class labels and objectness scores are also estimated. Among them, a bounding box having highest objectness score is selected and the remaining bounding boxes are removed.}
\label{fig:objectness}
\end{figure}

Two-class classification needs to be implemented in a given problem.
However, the estimation result may contain multiple estimated bounding boxes that correspond to multiple object classes for an image.
When multiple bounding boxes are estimated from an image, we perform non-maximal suppression \cite{yolo1} using objectness scores.
This process is illustrated in Fig. \ref{fig:objectness}.
Estimated bounding boxes having non-maximal objectness scores are removed.
After this process, we obtain one or zero estimated bounding box per image.
A class label corresponds to an estimated bounding box of an image is defined as the estimated class label of the image.

\section{Experiments} \label{sec:experiments}

We applied the proposed method to anterior eye images.
The images consist of 100 images of infective disease and 96 images of non-infective disease.
These images were taken at multiple medical institutions.
Ground truth class labels of the images were assigned by ophthalmologists.
Ground truth bounding boxes were given by an engineering researcher.

A Windows 10 PC with an NVIDIA TITAN V GPU was used to perform experiments.
The method was implemented using the Keras \cite{keras} with the TensorFlow library.

\section{Results}

Five fold cross validation was performed to evaluate classification performance of the proposed method.
In the result, 88.3\% of images (173/196) were correctly classified.
22 images were classified into wrong classes.
One image was not classified (no bounding box was estimated).
A confusion matrix of the classification result is shown in Table \ref{tab:confmatrix}.

\section{Discussion}

Our method successfully classified the anterior eye images into two classes with the high classification accuracy.
Automated classification of anterior eye images for infective and non-infective diseases have never been tackled before.
This is the first trial that tackles the classification problem and achieved the practical performance.
Our method can be used as an automated diagnosis assistance of eye diseases.
As shown in Table \ref{tab:confmatrix}, classification accuracies of infective and non-infective classes were similar.
89.0\% (89/100) of infective images were correctly classified.
Also, 87.5\% (84/96) of non-infective images were correctly classified.
This means the proposed classification method successfully extracted feature values that are useful to classify them from training images.
We addressed the image classification task as detection task of the corneas.
By using the bounding box annotations, the YOLOv3 focuses to distinguish the corneas.
Between the cases of infective and non-infective diseases, the appearances of the corneas are different.
Therefore, distinguishing the corneas based on its appearances is effective to the classification task of infective and non-infective cases.

We translated the anterior eye image classification task to the object detection task to make our method focuses on finding a specific anatomical structure that contributes to perform classification.
We selected the cornea as the object detection target because its appearance is different between cases of infective and non-infective diseases.
The ``anatomical structure focused image classification'' approach resulted in the high classification performance.

\begin{table}[tb]
\begin{center}
\caption{Confusion matrix of classification result}
\label{tab:confmatrix}
\begin{tabular}{|c|c|c|c|c|} 
\hline
\multicolumn{2}{|c|}{} & \multicolumn{3}{|c|}{Classification result} \\ \cline{3-5}
\multicolumn{2}{|c|}{} & Infective & Non-infective & Not classified \\ \hline
Ground truth & Infective & 89 & 10 & 1 \\ \cline{2-5}
 & Non-infective & 12 & 84 & 0 \\ \hline
\end{tabular}
\end{center}
\end{table}

\begin{figure}[tb]
\begin{center}
\begin{tabular}{c}
\includegraphics[width=0.95\textwidth, clip, trim=0 360 310 0]{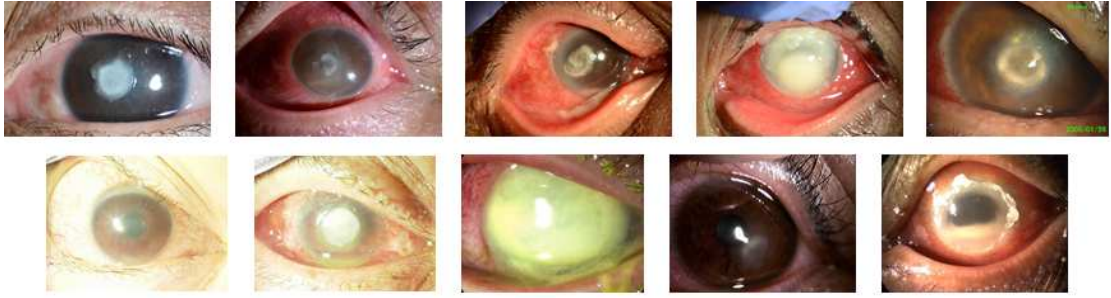} \\
(a) \\
\includegraphics[width=0.95\textwidth, clip, trim=0 460 340 0]{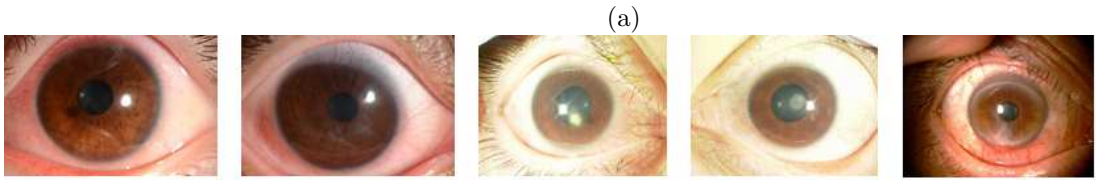} \\
(b)
\end{tabular}
\end{center}
\caption{Samples of classification results of infective disease images: (a) correctly classified infective disease images. (b) incorrectly classified infective disease images (classified as non-infective).}
\label{fig:result_infective}
\end{figure}

\begin{figure}[tb]
\begin{center}
\begin{tabular}{c}
\includegraphics[width=0.95\textwidth, clip, trim=0 360 245 0]{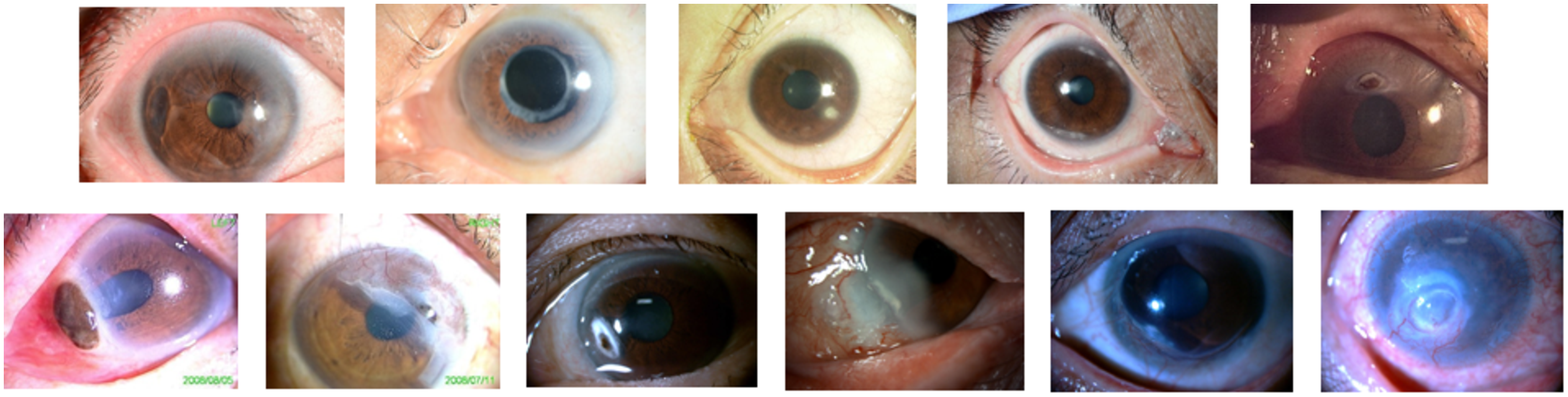} \\
(a) \\
\includegraphics[width=0.95\textwidth, clip, trim=0 460 245 0]{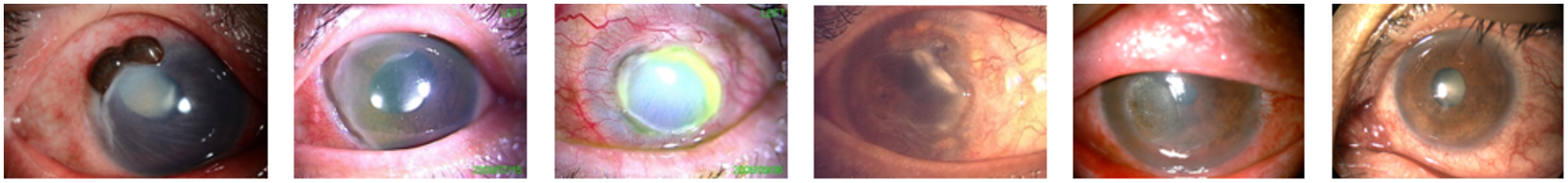} \\
(b)
\end{tabular}
\end{center}
\caption{Samples of classification results of non-infective disease images: (a) correctly classified non-infective disease images. (b) incorrectly classified non-infective disease images (classified as infective).}
\label{fig:result_noninfective}
\end{figure}

Figure \ref{fig:result_infective} shows infective disease images classified into infective and non-infective disease classes by the proposed method.
Also, non-infective disease images classified into non-infective and infective disease classes are shown in Fig. \ref{fig:result_noninfective}.
The proposed method was robust to differences of eye position, degree of open of the eyelid, and brightness of illumination.
Wrong classifications may be caused by specular on the corneas.
Position of opacified areas in the cornea is one criteria of differentiating infective and non-infective diseases.
Both of opacified areas and specular on the corneas are observed as white regions in images.
Therefore, miss-classifications may be caused by existence of specularity.
A removal process of specularity on the eye is necessary to improve classification accuracy.

\section{Conclusions}

We proposed a classification method of cases of infective and non-infective diseases from anterior eye images.
Because the anterior eye images have wide variation of their appearances, our method focuses on distinguishing the cornea.
The position of opacified area in the cornea is different among the two disease cases.
In our method, a object detection method finds infective cornea or non-infective cornea from the anterior eye images.
The two-class classification result of the image is made based on the object detection result.
In our experiment, the proposed method correctly classified 88.3\% of images.
The accuracy is enough high to be used as a diagnosis assistance of the anterior eye.
Future work includes classification of images into more detailed disease classes and utilization of other anatomical structure information.

\acknowledgments 
 
Parts of this research were supported by the AMED Grant Numbers 18lk1010028s0401,
19lk1010036h0001, and 19hs0110006h0003, the MEXT/JSPS KAKENHI Grant Numbers 26108006, 17H00867, and 17K20099, the JSPS Bilateral International Collaboration Grants.


\bibliography{20spie_paper_cite}   
\bibliographystyle{spiebib}   

\end{document}